\documentclass{article} 
\usepackage{iclr2026_conference,times}

\usepackage{amsmath,amsfonts,bm}

\def\eqref#1{equation~\ref{#1}}

\def\1{\bm{1}}

\DeclareMathAlphabet{\mathsfit}{\encodingdefault}{\sfdefault}{m}{sl}
\SetMathAlphabet{\mathsfit}{bold}{\encodingdefault}{\sfdefault}{bx}{n}

\usepackage[utf8]{inputenc} 
\usepackage[T1]{fontenc}    
\usepackage[hidelinks]{hyperref}  
\usepackage{url}            
\usepackage{booktabs}       
\usepackage{amsfonts}       
\usepackage{nicefrac}       
\usepackage{microtype}      
\usepackage{xcolor}         
\usepackage{graphicx}

\usepackage{amsmath}
\usepackage{amssymb}
\usepackage{mathtools}
\usepackage{amsthm}
\usepackage{makecell}
\usepackage{comment}
\usepackage[capitalize,noabbrev]{cleveref}

\usepackage{subcaption}
\usepackage{lipsum}
\usepackage{algorithm}
\usepackage{algpseudocode}
\usepackage{algorithmicx}
\usepackage{bm}
\usepackage{wrapfig}
\usepackage[super]{nth}
\usepackage{pifont}

\title{Beyond Objective Expressivity: Geometry Preservation in Multimodal Contrastive Learning}

\author{
  \textbf{Tillmann Rheude\textsuperscript{1,2$^\dagger$}}, 
  \textbf{Roland Eils\textsuperscript{1,2,3$^\dagger$}}, 
  \textbf{Benjamin Wild\textsuperscript{1$^\dagger$}}\\
  \textsuperscript{1}Berlin Institute of Health, Charité - Universitätsmedizin Berlin, 
  \textsuperscript{2}Department of Mathematics\\and Computer Science, Freie Universität Berlin, 
  \textsuperscript{3}Intelligent Medicine Institute, Fudan University\\
  $^\dagger$\texttt{\{tillmann.rheude, roland.eils, benjamin.wild\}@bih-charite.de}
}

\iclrfinalcopy

\usepackage[frozencache,cachedir=minted-cache]{minted}
\setminted{
    style=borland,
    linenos,
    numbersep=4pt,
    xleftmargin=1em,
}

\usepackage[acronym,toc,shortcuts]{glossaries}
\glsdisablehyper
\newacronym{ukb}{UKB}{UK Biobank}
\newacronym{sota}{sota}{state-of-the-art}
\newacronym{flops}{FLOPS}{Floating Point Operations}
\newacronym{ce}{CE}{Cross Entropy}
\newacronym{bce}{BCE}{Binary Cross Entropy}
\newacronym{ecg}{ECG}{electrocardiogram}
\newacronym{ehr}{EHR}{electronic health record}
\newacronym{mri}{MRI}{Magnetic Resonance Images}
\newacronym{ct}{CT}{computer-assisted tomography}
\newacronym{prs}{PRS}{polygenic risk score}
\newacronym{auroc}{AUROC}{area under the receiver operating characteristic}
\newacronym{vit}{ViT}{Vision Transformer}
\newacronym{mlp}{MLP}{Multi-Layer Perceptron}
\newacronym{cnn}{CNN}{Convolutional Neural Network}
\newacronym{sd}{SD}{standard deviation}
\newacronym{se}{SE}{standard error}
\newacronym{gpu}{GPU}{Graphics Processing Unit}
\newacronym{mip}{MIP}{multilinear inner product}
\newacronym{tc}{TC}{total correlation}
\newacronym{moe}{MoE}{mixture-of-experts model}
\newacronym{poe}{PoE}{product-of-experts model}
\newacronym{ood}{OOD}{Out-of-Distribution}
\newacronym{cam}{CAM}{class activation map}
\newacronym{gpe}{GPE}{geometry-preserving encoder}
\newacronym{hpc}{HPC}{high-performance cluster}
\newacronym{icu}{ICU}{intensive care unit}
\newacronym{jvp}{JVP}{Jacobian-vector product}
\newacronym{vlm}{VLM}{vision-language model}

\newcommand{\ie}{\textit{, i.e., }}
\newcommand{\eg}{\textit{, e.g., }}

\begin{document}

\maketitle

\begin{abstract}
    Contrastive learning is increasingly moving toward settings with three or more modalities instead of image-text pairs.
    Yet, extending models from pairwise to higher-order multimodal alignment can introduce optimization and representation challenges.
    We identify encoder Jacobian conditioning as a key factor in trimodal contrastive learning: poorly conditioned encoders exhibit collapsing or amplified singular-value spectra, leading to exploding Jacobian condition numbers and degraded multimodal alignment.
    We introduce \acrfullpl{gpe} by directly conditioning the Jacobian through regularization and demonstrating that simple modifications like LeakyReLU activations and residual paths recover these geometric benefits.
    Across a synthetic benchmark and four real-world datasets including missing modalities, improving Jacobian conditioning boosts retrieval and linear probe performance across multiple contrastive objectives, whereas expressive objectives yield little benefit in linear probes.
    More broadly, our results show that multimodal contrastive learning depends not only on objective expressivity, but also on the geometric and optimization properties of the underlying encoders.\footnote{The code repository is available on \href{https://github.com/TillmannRheude/gpe}{GitHub}.}
\end{abstract}

\section{Introduction}
\label{sec:introduction}

\begin{wrapfigure}[15]{r}{0.5\textwidth}
        \vspace{-.52cm}
        \centering
        \includegraphics[width=\linewidth]{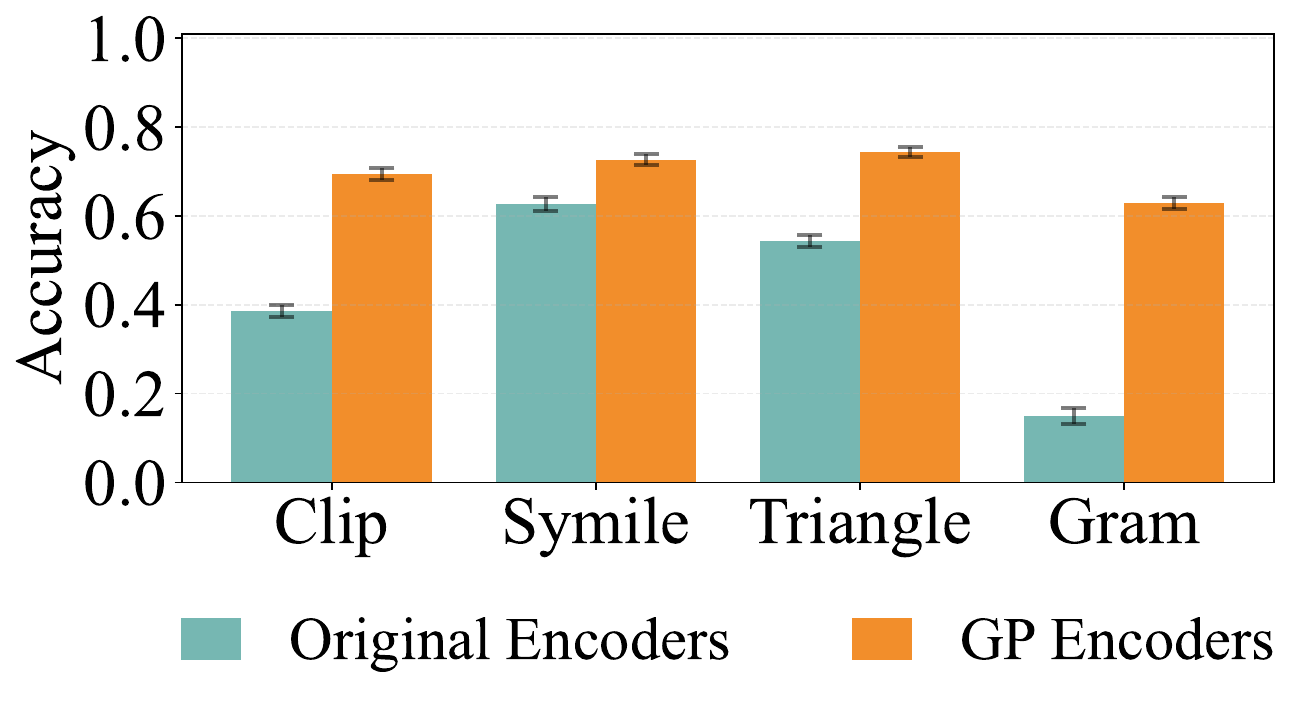}
        \caption{
            Well-tuned standard encoders vs. \acsp{gpe} across multiple multimodal contrastive objectives on the \acs{ukb}. \acsp{gpe} improve retrieval performance for all evaluated contrastive objectives.
        }
        \label{fig:gp_vs_original}
\end{wrapfigure}

Multimodal contrastive learning has emerged as a powerful method for aligning representations across modalities. While early work focused on bimodal \acp{vlm} \citep{Radford_Kim_Hallacy_Ramesh_Goh_Agarwal_Sastry_Askell_Mishkin_Clark_2021}, many real-world applications involve more modalities. 
For example, in life sciences, modalities may cover demographics, clinical timeseries, laboratory measurements, free-text reports, and neural or behavioral recordings \citep{Acosta_Falcone_Rajpurkar_Topol_2022,Schneider_Lee_Mathis_2023}.
Motivated by these settings, recent work has proposed higher-order contrastive objectives that move beyond pairwise alignment \citep{Saporta_Puli_Goldstein_Ranganath_2024,Cicchetti_Grassucci_Comminiello_2025,cicchetti2025gramian,Dufumier_Navarro_Tuia_Thiran_2025}.
At the same time, multimodal learning relies on heterogeneous encoder families, ranging from ResNets \citep{He_Zhang_Ren_Sun_2016} and Transformers \citep{Vaswani_Shazeer_Parmar_Uszkoreit_Jones_Gomez_Kaiser_Polosukhin_2017} to \ac{mlp}-based encoders \citep{Rosenblatt_1957}, which are frequently used to process tabular modalities and map pretrained representations into shared embedding spaces.
Different encoder families exhibit different optimization characteristics, and some encoders such as \acp{mlp} may be more sensitive to optimization pathologies, as highlighted in \Cref{fig:gp_vs_original}. This raises a complementary question to objective design: how do the geometric properties of modality encoders affect multimodal contrastive learning?
\begin{figure}[t]
    \centering
    \includegraphics[width=\linewidth]{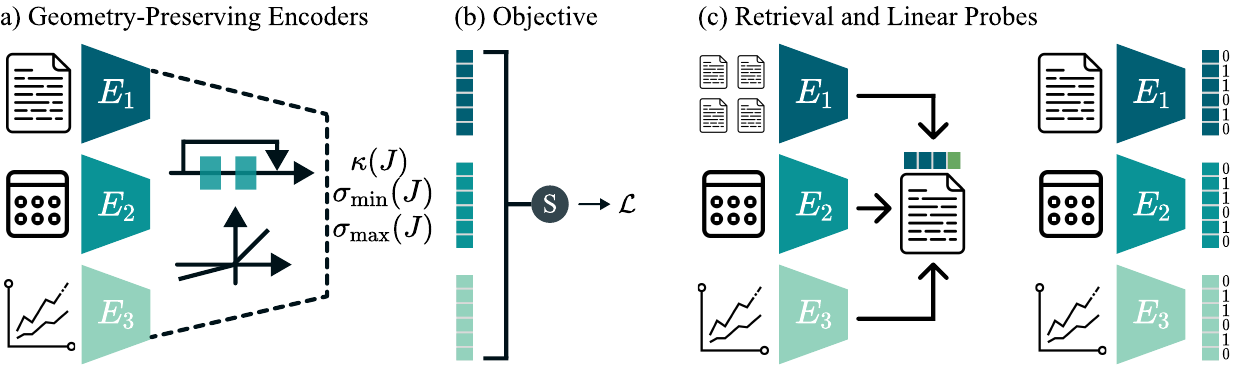}
    \caption{
        Multimodal contrastive learning with \acsp{gpe} exemplified for MIMIC-IV \citep{Johnson_Bulgarelli_Shen_Gayles_Shammout_Horng_Pollard_Hao_Moody_Gow_etal_2023,Johnson_Bulgarelli_Pollard_Horng_Celi_Mark}. Radiology reports, demographics, and timeseries are processed by LoRA-finetuned RadBERT \citep{Yan_McAuley_Lu_Du_Chang_Gentili_Hsu_2022} and \ac{mlp} encoders $E_i$.
        \textbf{(a)} \acsp{gpe} maintain well-conditioned Jacobians through residual transport paths, and LeakyReLU activations, preventing the condition number $\kappa$ defined by singular values $\sigma_{\min}$ and $\sigma_{\max}$ to explode. 
        \textbf{(b)} \acsp{gpe} can be combined with arbitrary multimodal contrastive objectives $S$, such as Clip. 
        \textbf{(c)} The resulting models and representations are extensively tuned and evaluated on retrieval and downstream linear probing tasks. Our results demonstrate that preserving encoder geometry improves training dynamics, retrieval performance, and downstream linear probes across diverse multimodal contrastive objectives and datasets.
    }
    \label{fig:overview}
\end{figure}
\newpage In contrast to the growing multimodal contrastive learning literature on objective design and representation geometry \citep{Wang_Isola_2020,Liang_Zhang_Kwon_Yeung_Zou_2022,Jing_Vincent_LeCun_Tian_2022,Yi_Douady_Chen_2025,Cai_Zhang_Liu_Shi_2026}, relatively little attention has been paid to the geometry of the encoders that produce these representations \citep{Golovanevsky_Mahableshwarkar_Eickhoff_Singh_2025,Rheude_Hegselmann_Eils_Wild_2026}. Meanwhile, a large body of work has linked neural network trainability to Jacobian conditioning and dynamical isometry, showing that poorly conditioned Jacobians impair signal propagation, optimization, and representation learning \citep{Saxe_McClelland_Ganguli_2014,Pennington_Schoenholz_Ganguli_2017,Burkholz_Dubatovka_2019,Xiao_Bahri_Sohl_Dickstein_Schoenholz_Pennington_2018,Tarnowski_Warchol_Jastrzcebski_Tabor_Nowak_2019}. 
We show that similar phenomena arise in multimodal contrastive learning: Alignment difficulties are characterized by exploding Jacobian condition numbers. 

In this paper, we study encoder geometry in trimodal contrastive learning. 
We theoretically and empirically connect multimodal contrastive learning to ill-conditioned encoder Jacobians.
This is particularly relevant when heterogeneous modalities are used. For example, tabular modalities might be encoded by \acp{mlp}, while radiology reports might be represented by a LoRA-finetuned BERT model followed by an \ac{mlp} projection head (\Cref{fig:overview}).
First, we test whether encoder Jacobian conditioning has a mechanistic effect on multimodal contrastive learning through Jacobian regularization.
Then, we introduce \acp{gpe}, a family of lightweight architectural interventions based on residual paths \citep{He_Zhang_Ren_Sun_2016} and LeakyReLU activations \citep{Maas_Hannun_Ng_2013} that provide similar geometric benefits at lower computational cost. 
Across a controlled synthetic benchmark \citep{Rheude_Hegselmann_Eils_Wild_2026}, \ac{ukb} cohorts with complete and missing modalities \citep{Sudlow_Gallacher_Allen_Beral_Burton_Danesh_Downey_Elliott_Green_Landray_etal_2015}, and the MIMIC-IV benchmark \citep{Johnson_Bulgarelli_Shen_Gayles_Shammout_Horng_Pollard_Hao_Moody_Gow_etal_2023,Johnson_Bulgarelli_Pollard_Horng_Celi_Mark}, we show that preserving encoder geometry improves optimization stability, retrieval performance, and downstream linear probing for diverse contrastive objectives. 
Our results suggest that objective expressivity alone is insufficient: while increasingly expressive objectives primarily affect retrieval performance, preserving encoder geometry improves retrieval and downstream linear probes across objectives. 
Our contributions are summarized as follows:
\begin{itemize}
    \item We identify encoder Jacobian conditioning as a key factor in multimodal contrastive learning and show that it influences the trainability of \ac{mlp}-based modality encoders.

    \item We introduce \acp{gpe}\ie practical interventions based on residual paths and LeakyReLU activations that improve optimization stability, and retrieval and linear probe performance.

    \item We demonstrate that ill-conditioned encoders appear across multiple contrastive objectives on both synthetic and real-world datasets including missing modalities.
    
    \item We show that \acp{gpe} consistently improve retrieval and downstream linear probes, whereas increasingly expressive multimodal objectives primarily improve retrieval.
\end{itemize}

\section{Related Work}
\label{sec:related_work}

\paragraph{Multimodal Contrastive Learning}
Multimodal contrastive learning has become a central approach for learning representations across modalities with\eg \acp{vlm} such as Clip \citep{Radford_Kim_Hallacy_Ramesh_Goh_Agarwal_Sastry_Askell_Mishkin_Clark_2021} and Align \citep{Jia_Yang_Xia_Chen_Parekh_Pham_Le_Sung_Li_Duerig_2021}. Subsequent work improved scalability and training formulations through approaches such as LiT \citep{Zhai_Wang_Mustafa_Steiner_Keysers_Kolesnikov_Beyer_2022} and SigLIP \citep{Zhai_Mustafa_Kolesnikov_Beyer_2023}. More recent methods move beyond pairwise alignment by introducing higher-order or many-modal objectives, including Symile \citep{Saporta_Puli_Goldstein_Ranganath_2024}, Triangle \citep{Cicchetti_Grassucci_Comminiello_2025}, CoMM \citep{Dufumier_Navarro_Tuia_Thiran_2025}, ConFu \citep{Koutoupis_Zervou_Kontras_DeVos_Tsakalides_Tsagkatakis_2026}, and Gram \citep{cicchetti2025gramian}.
Further, relatively little attention has been paid to the geometry and optimization properties of the encoders themselves with approaches like Gated Symile \citep{Rheude_Hegselmann_Eils_Wild_2026} and PiCME \citep{Golovanevsky_Mahableshwarkar_Eickhoff_Singh_2025} highlighting the closest related work. 
In contrast, we study encoder Jacobian conditioning across multimodal contrastive objectives, evaluate both retrieval and linear probes, and show that expressive objectives can still underperform when the encoders producing the representations become poorly conditioned.

\paragraph{Geometry, Dynamical Isometry, and Signal Propagation}
Neural network optimization is commonly linked to the singular value spectrum of network Jacobians. 
Dynamical isometry studies show that networks train more reliably when Jacobian spectra remain well-conditioned throughout depth \citep{Saxe_McClelland_Ganguli_2014,Pennington_Schoenholz_Ganguli_2017}. 
Related work has investigated how initialization schemes can promote favorable Jacobian spectra and dynamical isometry \citep{Pennington_Schoenholz_Ganguli_2017}, how activation functions influence signal propagation and trainability \citep{Burkholz_Dubatovka_2019}, how Jacobian regularization can improve local representation stability \citep{Rifai_Vincent_Muller_Glorot_Bengio_2011}, how network Jacobians relate to robustness and generalization \citep{Sokolic_Giryes_Sapiro_Rodrigues_2017}, and how architectural design choices affect optimization dynamics \citep{Hoffman_Roberts_Yaida_2019}. 
Further, Jacobian spectra have been connected to feature learning dynamics \citep{Xiao_Bahri_Sohl_Dickstein_Schoenholz_Pennington_2018}, the effectiveness of Jacobian regularization \citep{Xue_Whitecross_Mirzasoleiman_2022}, the role of data structure and activation functions in neural network learning dynamics \citep{Sonthalia_Murray_Montufar_2026}, and alignment stability in multimodal models \citep{Garg_Saratchandran_Lucey_2026}.
Separately, contrastive learning research has studied alignment and uniformity of learned representations \citep{Wang_Isola_2020}, dimensional collapse in contrastive embeddings \citep{Jing_Vincent_LeCun_Tian_2022}, modality-gap formation and multimodal representation structure \citep{Liang_Zhang_Kwon_Yeung_Zou_2022,Yi_Douady_Chen_2025}, alignment under missing modalities \citep{Poklukar_Vasco_Yin_Melo_Paiva_2022}, and geometric analyses of contrastive objectives \citep{Cai_Zhang_Liu_Shi_2026}.
In contrast, we study the conditioning of modality-specific encoder Jacobians in multimodal contrastive learning and show its impact on both retrieval and downstream representation quality.

\paragraph{Residual Architectures and Optimization Stability}
Residual connections are central to modern deep architectures because they provide identity transport paths that stabilize optimization and gradient flow \citep{He_Zhang_Ren_Sun_2016}. 
Correspondingly, prior work has shown that residual networks can preserve favorable input-output Jacobian spectra and achieve dynamical isometry across a broad range of activation functions \citep{Tarnowski_Warchol_Jastrzcebski_Tabor_Nowak_2019}. 
This principle underlies residual architectures such as Transformer architectures built around skip connections \citep{Vaswani_Shazeer_Parmar_Uszkoreit_Jones_Gomez_Kaiser_Polosukhin_2017}, and residual \acp{mlp} such as ResMLPs \citep{Touvron_Bojanowski_Caron_Cord_El_Nouby_Grave_Izacard_Joulin_Synnaeve_Verbeek_et_al_2023}. Related stabilization methods include learnable residual gating through ReZero \citep{Bachlechner_Majumder_Mao_Cottrell_McAuley_2021}, initialization schemes such as Fixup \citep{Huang_Perez_Ba_Volkovs_2020}, and normalization strategies such as DeepNorm \citep{Wang_Ma_Dong_Huang_Zhang_Wei_2024}.
More recently, \citet{Ji_Saratchandran_Moghadam_Lucey_2025} argue that skip connections are a primary driver of Transformer trainability, potentially more important than the expressivity of the attention blocks themselves. 
Complementing this line of work, we study how encoder conditioning affects multimodal contrastive learning. 
Through direct Jacobian regularization, we provide interventional evidence that improving encoder conditioning stabilizes optimization and demonstrate that residual transport provides a practical and computationally efficient mechanism for achieving similar geometric benefits.
\section{Method}
\label{sec:method}

\subsection{Multimodal Contrastive Learning}

We consider trimodal observations 
\(
(x_1,x_2,x_3)
\)
analogous to prior work on multimodal contrastive learning (\Cref{sec:related_work}) and modality-specific encoders
\begin{equation}
    E_i:\mathcal{X}_i \rightarrow \mathbb{R}^d,
    \qquad
    i \in \{1,2,3\}.
\end{equation}
The resulting modality representations are $z_i = E_i(x_i)$. A multimodal contrastive objective assigns a compatibility score $s(z_1,z_2,z_3)$ to aligned triples and contrasts it against mismatched triples sampled from the batch. Most existing work focuses on increasing the expressivity of the scoring function \(s\). In contrast, we investigate a complementary question: how does the geometry of the modality encoders $E_i$ influence whether a multimodal contrastive objective can be optimized successfully?

\subsection{Encoder Geometry and Jacobian Conditioning}
The geometry induced by encoder \(E_i\) is defined by its input-output Jacobian
\begin{equation}
    J_i(x)
    =
    \frac{\partial E_i(x)}{\partial x}.
\end{equation}
Let
\(
\sigma_{\min}(J_i)
\)
and
\(
\sigma_{\max}(J_i)
\)
denote the smallest and largest singular values of \(J_i\), respectively. We quantify encoder conditioning through the Jacobian condition number
\begin{equation}
    \kappa(J_i)
    =
    \frac{\sigma_{\max}(J_i)}
         {\sigma_{\min}(J_i)}.
\end{equation}
For a small perturbation \(\delta\),
\begin{equation}
    E_i(x+\delta)-E_i(x)
    \approx
    J_i(x)\delta.
\end{equation}
The extremal singular values characterize the local amplification of perturbations:
\begin{equation}
    \sigma_{\min}(J_i(x))\|\delta\|_2
    \le
    \|J_i(x)\delta\|_2
    \le
    \sigma_{\max}(J_i(x))\|\delta\|_2.
\end{equation}
Small values of \(\sigma_{\min}\) indicate that some input directions are strongly suppressed, whereas large values of \(\sigma_{\max}\) indicate excessive amplification.
Both effects degrade Jacobian conditioning and can impair information and gradient transport.
Moreover, gradients from the contrastive objective with loss $\mathcal{L}$ propagate according to
\begin{equation}
    \nabla_x \mathcal{L}
    =
    J_i(x)^\top \nabla_{z_i}\mathcal{L},
\end{equation}
showing that poorly conditioned Jacobians directly impair gradient transport from the multimodal objective back to the modality-specific inputs.
For example, for an \ac{mlp} with $L$ layers, weights $W_\ell$, ReLU activations, and diagonal activation Jacobians \(D_\ell\), the encoder Jacobian can be written as
\begin{equation}
    J_i(x)
    =
    W_L D_{L-1} W_{L-1}
    \cdots
    D_1 W_1.
    \label{eq:jacobian_definition_mlp}
\end{equation}
Since ReLU activation Jacobians contain only zeros and ones, entire directions can be removed during forward and backward propagation, leading to rank loss and collapsing $\sigma_{\min}$. Conversely, products of weight matrices may excessively amplify certain directions, resulting in large $\sigma_{\max}$.

Interestingly, multimodal contrastive learning requires encoders to preserve modality-specific information rather than immediately transform it. However, learning near-identity mappings is itself challenging for plain ReLU \acp{mlp}, whose Jacobians are products of weights and activation gates (\Cref{eq:jacobian_definition_mlp}) \citep{He_Zhang_Ren_Sun_2016}.

\subsection{Jacobian Intervention}

The analysis above suggests that encoder Jacobian conditioning may play a mechanistic role in multimodal contrastive learning. To test this hypothesis, we directly intervene on the encoder Jacobian through stochastic Jacobian regularization.
Computing singular values of \(J_i\) exactly is computationally expensive for high-dimensional encoders. Instead, we employ an approximation based on \acp{jvp}. For a random direction \(v\), we define the directional gain
\begin{equation}
    r_i(x,v)
    =
    \frac{\|J_i(x)v\|_2^2}
         {\|v\|_2^2}.
\end{equation}

Given \(K\) sampled directions, we estimate
\begin{equation}
    r_{\min}(x)
    =
    \min_{k \le K}
    r_i(x,v_k),
    \qquad
    r_{\max}(x)
    =
    \max_{k \le K}
    r_i(x,v_k).
\end{equation}

We then penalize deviations from a desired directional-gain band \([m,M]\):
\begin{equation}
    \mathcal{L}_{\mathrm{JVP}}
    =
    w_{\min}
    \underbrace{
    \Big[
    \max(0,m^2-r_{\min}(x))
    \Big]^2
    }_{\substack{\text{prevent directional collapse}}}
    + 
    w_{\max}
    \underbrace{
    \Big[
    \max(0,r_{\max}(x)-M^2)
    \Big]^2
    }_{\substack{\text{prevent directional explosion}}}.
\end{equation}
Importantly, \(r_{\min}\) and \(r_{\max}\) are not estimates of the extremal singular values themselves. Rather, they summarize directional gains observed along randomly sampled directions and bias optimization toward better-conditioned Jacobians.
Consequently, improvements obtained through Jacobian regularization provide evidence that encoder conditioning influences multimodal alignment performance rather than merely correlating with it.
While effective, Jacobian regularization introduces additional computational overhead through repeated \acp{jvp}. We therefore seek lightweight architectural mechanisms that naturally preserve encoder geometry throughout training.

\subsection{Geometry-Preserving Encoders}
We introduce \acrfullpl{gpe}\ie encoder modifications to maintain well-conditioned Jacobians throughout contrastive optimization. 
We find that two surprisingly simple modifications\ie residual transport paths and LeakyReLU activations recover many of the benefits obtained through direct Jacobian intervention.

\paragraph{Residual Transport}
We introduce residual transport paths around modality-specific encoders. Given a nonlinear branch \(h_i\), we define
\begin{equation}
    E_i(x)
    =
    P_i x + h_i(x),
\end{equation}
where \(P_i\) is a learned projection. The corresponding Jacobian is
\begin{equation}
    J_{E_i}(x)
    =
    P_i + J_{h_i}(x).
\end{equation}
By standard singular-value perturbation bounds (\Cref{app:residual_theorem}),
\begin{equation}
    \sigma_{\min}(J_{E_i})
    \ge
    \sigma_{\min}(P_i)
    -
    \|J_{h_i}\|_2, 
    \hspace{15pt} \text{and} \hspace{15pt} 
    \sigma_{\max}(J_{E_i})
    \le
    \sigma_{\max}(P_i)
    +
    \|J_{h_i}\|_2.
\end{equation}
Consequently, if $\|J_{h_i}\|_2 < \sigma_{\min}(P_i)$, then $\sigma_{\min}(J_{E_i}) > 0$.
Thus, a well-conditioned projection \(P_i\) provides an additive transport path that preserves Jacobian conditioning by simultaneously supporting weak directions and limiting excessive amplification.

\paragraph{LeakyReLU Activations}
We replace ReLU with LeakyReLU activations \citep{Maas_Hannun_Ng_2013}
\begin{equation}
\phi_\alpha(u)
=
\begin{cases}
u, & u > 0, \\
\alpha u, & u \le 0,
\end{cases}
\end{equation}
where the negative slope \(\alpha > 0\).
Unlike ReLU, whose derivative can become exactly zero,
\begin{equation}
    \frac{\partial \phi_\alpha(u)}{\partial u}
    \in
    \{\alpha,1\}.
\end{equation}
for \(\alpha>0\). Consequently, the activation Jacobian
\(
D_\ell
\)
satisfies
\begin{equation}
    \alpha
    \le
    \sigma_{\min}(D_\ell)
    \le
    \sigma_{\max}(D_\ell)
    \le
    1.
\end{equation}
Unlike ReLU, LeakyReLU prevents activation derivatives from becoming exactly zero. Consequently, activation gates can no longer completely remove directions from the encoder Jacobian. While rank loss may still arise from weights or architectural bottlenecks, LeakyReLU removes an important source of singular-value collapse.
\section{Experiments}
\label{sec:experiments}

\subsection{Datasets}
\label{sec:datasets}
To assess the proposed \acp{gpe} in the multimodal contrastive setting, we consider a diverse set of datasets (\Cref{tab:dataset_summary}). 
Synthetic-\texttt{XNOR} \citep{Rheude_Hegselmann_Eils_Wild_2026} provides a controlled environment for studying higher-order multimodal alignment and optimization dynamics. MIMIC-IV \citep{Johnson_Bulgarelli_Shen_Gayles_Shammout_Horng_Pollard_Hao_Moody_Gow_etal_2023,Johnson_Bulgarelli_Pollard_Horng_Celi_Mark} and MIMIC-Symile \citep{Saporta_Puli_Goldstein_Ranganath_2024} serves as a clinically relevant benchmark combining heterogeneous modalities. Finally, the \ac{ukb} \citep{Sudlow_Gallacher_Allen_Beral_Burton_Danesh_Downey_Elliott_Green_Landray_etal_2015} allows us to evaluate performance at population scale, while \ac{ukb}-Union additionally introduces modality missingness.

\paragraph{Synthetic-XNOR}
We adopt Synthetic-\texttt{XNOR} \citep{Rheude_Hegselmann_Eils_Wild_2026}, a controlled trimodal benchmark designed to expose failure modes of higher-order multimodal contrastive objectives. Rather than requiring all modalities to contribute equally, the task tests whether an objective can combine information selectively, including cases where one modality is weakly informative, misaligned, or should be effectively ignored for successful retrieval. This makes the benchmark useful for studying optimization dynamics and alignment failures in a controlled setting, isolated from real-world confounders. We use Synthetic-\texttt{XNOR} to analyze how encoder Jacobian conditioning and especially direct Jacobian regularization interacts with higher-order contrastive objectives.

\paragraph{MIMIC}
We construct a benchmark from MIMIC-IV \citep{Johnson_Bulgarelli_Shen_Gayles_Shammout_Horng_Pollard_Hao_Moody_Gow_etal_2023,Johnson_Bulgarelli_Pollard_Horng_Celi_Mark,Goldberger_Amaral_Glass_Hausdorff_Ivanov_Mark_Mietus_Moody_Peng_Stanley_2000} with radiology reports, clinical timeseries (\Cref{app:additional_details}), and demographics recorded during the first $48$ hours of admission. Radiology reports are encoded using a LoRA-finetuned RadBERT model \citep{Hu_shen_Wallis_Allen_Zhu_Li_Wang_Wang_Chen_2022,Yan_McAuley_Lu_Du_Chang_Gentili_Hsu_2022} followed by an \ac{mlp}. Timeseries and demographic variables are encoded using an \ac{mlp}.
Further, we use MIMIC-Symile \citep{Saporta_Puli_Goldstein_Ranganath_2024} which includes chest X-rays, \acp{ecg}, and laboratory values. Chest X-rays and \acp{ecg} are encoded with ResNets and laboratory values with an \ac{mlp} \citep{Saporta_Puli_Goldstein_Ranganath_2024}. We ablate them by removing the skip connections and comparing the resulting encoders against the standard variants (\Cref{fig:mimic_symile_loss_sd_comparison}).

\paragraph{\ac{ukb}}
We use the \ac{ukb} \citep{Sudlow_Gallacher_Allen_Beral_Burton_Danesh_Downey_Elliott_Green_Landray_etal_2015}, comprising proteomics, metabolomics, and \acp{ehr} \citep{Rheude_Eils_Wild_2025}. For the \ac{ehr} modality, we employ Qwen3-Embedding-8B embeddings \citep{Zhang_Li_Long_Zhang_Lin_Yang_Xie_Yang_Liu_Lin_etal_2025,Hegselmann_Arnim_Rheude_Kronenberg_Sontag_Hindricks_Eils_Wild_2025}, and all modality encoders are implemented as \acp{mlp}. We evaluate two complementary data-driven settings. First, \ac{ukb}-Intersection contains only samples with complete observations across all modalities (\(37{,}888\) samples). Second, \ac{ukb}-Union contains \(486{,}400\) samples and introduces substantial modality missingness. Following \citet{Saporta_Puli_Goldstein_Ranganath_2024}, missing modalities are represented through binary missingness masks appended to the corresponding inputs. The \ac{ukb} experiments enable evaluation of \acp{gpe} at population scale, both with complete multimodal observations and under realistic missing-modality conditions.

\begin{table*}[t]
    \centering
    \caption{
        Overview of benchmark datasets for our evaluation. Retrieval modality is listed first.
    }
    \label{tab:dataset_summary}
    \small
    \begin{tabular}{@{}l c c c@{}}
    \toprule
    \textbf{Dataset} & 
    \textbf{\# Samples} & 
    \textbf{Modalities} &
    \textbf{Linear Probe} \\
    \midrule

    Synthetic-\texttt{XNOR} \citep{Rheude_Hegselmann_Eils_Wild_2026}
        & $30,000$ & $A$, $B$, $C$ & n/a \\

    

    MIMIC-Symile \citep{Saporta_Puli_Goldstein_Ranganath_2024}
        & $10,345$ & Chest X-rays, \acp{ecg}, laboratory & n/a \\

    MIMIC-IV \citep{Johnson_Bulgarelli_Shen_Gayles_Shammout_Horng_Pollard_Hao_Moody_Gow_etal_2023,Johnson_Bulgarelli_Pollard_Horng_Celi_Mark}
        & $52,802$ & Reports, Clinicals, Demographics & Mortality \\

    \acs{ukb}-Intersection \citep{Sudlow_Gallacher_Allen_Beral_Burton_Danesh_Downey_Elliott_Green_Landray_etal_2015}
        & $37,888$
        & Proteomics, Metabolomics, \acs{ehr} & Mortality  \\

    \acs{ukb}-Union \citep{Sudlow_Gallacher_Allen_Beral_Burton_Danesh_Downey_Elliott_Green_Landray_etal_2015}
        & $486,400$
        & Proteomics, Metabolomics, \acs{ehr} & Mortality  \\

    \bottomrule
\end{tabular}
\end{table*}

\subsection{Experimental Setup}
\label{sec:experimental_setup}
All methods are implemented within a unified framework \citep{Falcon_Lightning_2019} to ensure fair comparisons across objectives, encoders, and datasets.
We follow best-practices for multimodal learning \citep{Rheude_Eils_Wild_2026} including extensive hyperparameter tuning (\Cref{app:hyperparameter_tuning}). 
We construct \acp{gpe} using both residual paths and LeakyReLU activations.
Further, we adopt the use of a learnable logit scale, and a pair-sampling strategy \citep{Rheude_Hegselmann_Eils_Wild_2026} for all objectives if applicable. Optimization is performed using ScheduleFree-AdamW \citep{Defazio_Yang_Khaled_Mishchenko_Mehta_Cutkosky_2024} with norm-based gradient clipping, helping distinguish Jacobian-induced optimization difficulties from instabilities caused by large gradients.
LoRA adapters are applied to the query and value projections of the Transformer attention layers. 
Tabular results on real-world datasets are evaluated using 5-fold cross-validation with mutually exclusive sample identifiers across folds. For each fold, models are trained with three independent random seeds and reported metrics are averaged across folds and seeds (mean $\pm$ \ac{se}).
Jacobian statistics for the training dynamics are computed separately for each modality encoder and then averaged across modalities.
Based on preliminary scaling experiments, we fix the batch size to \(128\) for Synthetic-\texttt{XNOR}, \(280\) for MIMIC-Symile, and \(512\) for MIMIC-IV and the \ac{ukb}.
We evaluate all methods in retrieval tasks and linear probes to assess whether improvements from objectives and \acp{gpe} translate to downstream performance, which is often more relevant than retrieval in life science applications.

\subsection{Jacobian Analyses}
\label{sec:jacobian_analyses}

\begin{figure}[t]
    \centering
    \includegraphics[width=\linewidth]{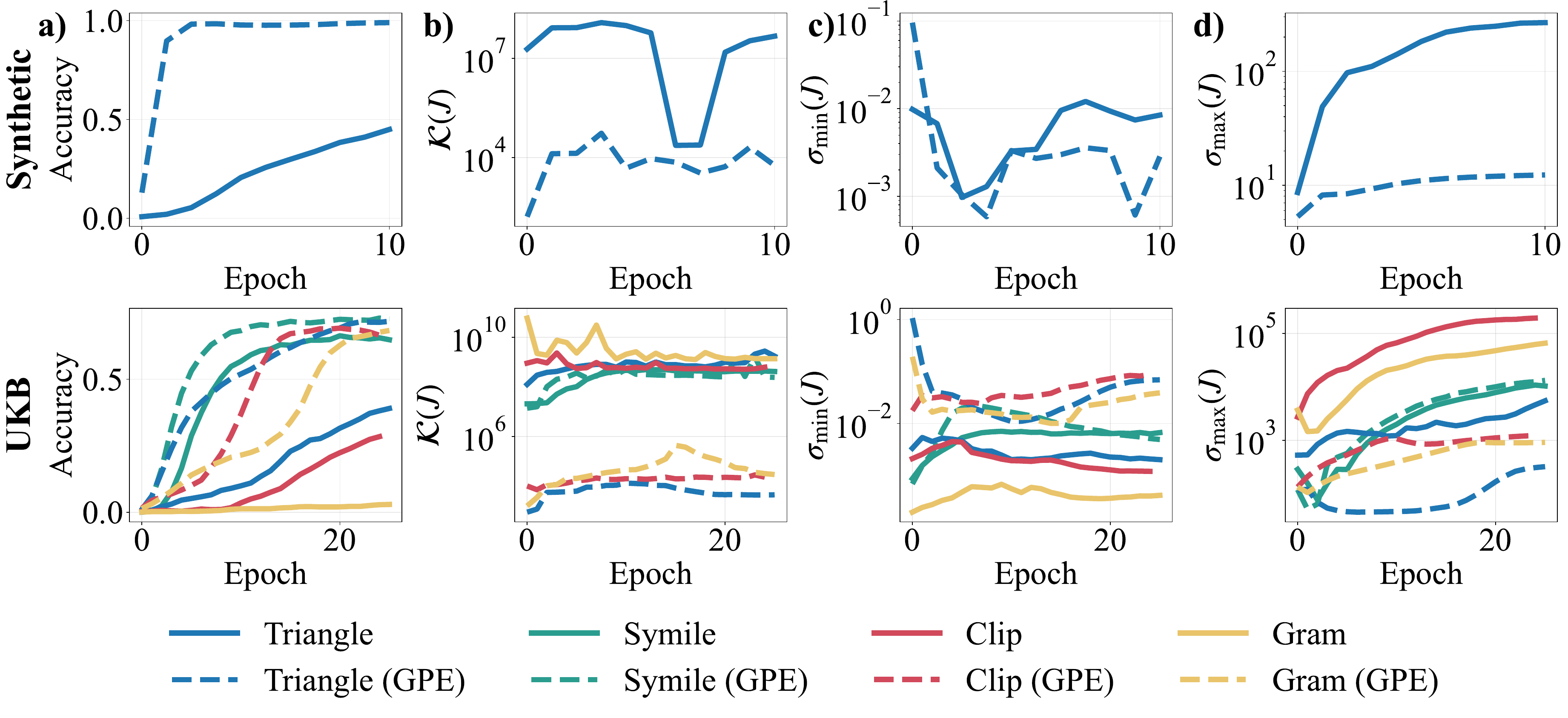}
    \caption{
        Training dynamics of well-tuned standard encoders and \acp{gpe}.
        The upper row focuses on Triangle and Synthetic-\texttt{XNOR} to highlight the effect of \acp{gpe}, whereas the lower row demonstrates their transferability to a real-world dataset, namely the \ac{ukb}.
        Standard \ac{mlp} encoders exhibit severe geometric degeneration, characterized by 
        \textbf{(a)} low accuracy, 
        \textbf{(b)} exploding condition numbers $\kappa(J)$, 
        \textbf{(c)} collapsing minimum singular values $\sigma_{\min}(J)$, and 
        \textbf{(d)} rapidly increasing maximum singular values $\sigma_{\max}(J)$. 
        In contrast, \acp{gpe} maintain performance and well-conditioned Jacobians throughout training. This demonstrates that objectives alone are insufficient for multimodal contrastive learning. 
    }
    \label{fig:epoch_vs_jacobian}
\end{figure}

\paragraph{Cross-Objective Training Dynamics}
Having established Jacobian conditioning as a potential source of optimization instability, we next examine whether this reflects a broader encoder-level limitation. We therefore compare training dynamics across objectives and datasets.
Despite their different scoring functions, standard \ac{mlp} encoders exhibit a general behavior (\Cref{fig:epoch_vs_jacobian}): retrieval accuracy is lower, $\kappa (J)$ rapidly increase, $\sigma_{\min}(J)$ collapse, and $\sigma_{\max}(J)$ grow. 
In contrast, \acp{gpe} generally maintain better-conditioned Jacobians and achieve both faster convergence and higher retrieval accuracies.
Interestingly, Symile on the \ac{ukb} constitutes a notable exception, exhibiting comparatively stable optimization even without \acp{gpe}. This behavior is not observed for Symile on Synthetic-\texttt{XNOR},  MIMIC-IV, and MIMIC-Symile (\Cref{app:additional_details,fig:synth_xnor_epoch_vs_jacobian,fig:mimic_iv_epoch_vs_jacobian,fig:mimic_symile_epoch_vs_jacobian}), suggesting that the interaction between objective design and encoder geometry may depend on datasets.
In general, these results indicate that objective expressivity alone is insufficient for successful multimodal contrastive learning and suggest that encoder geometry is a shared bottleneck across objectives.

\paragraph{Jacobian Regularization}
\begin{wrapfigure}[14]{r}{0.55\textwidth}
    \centering
    \vspace{-15pt}
    \includegraphics[
            width=\linewidth,
            trim=0 0.8cm 0 0,
            clip
    ]{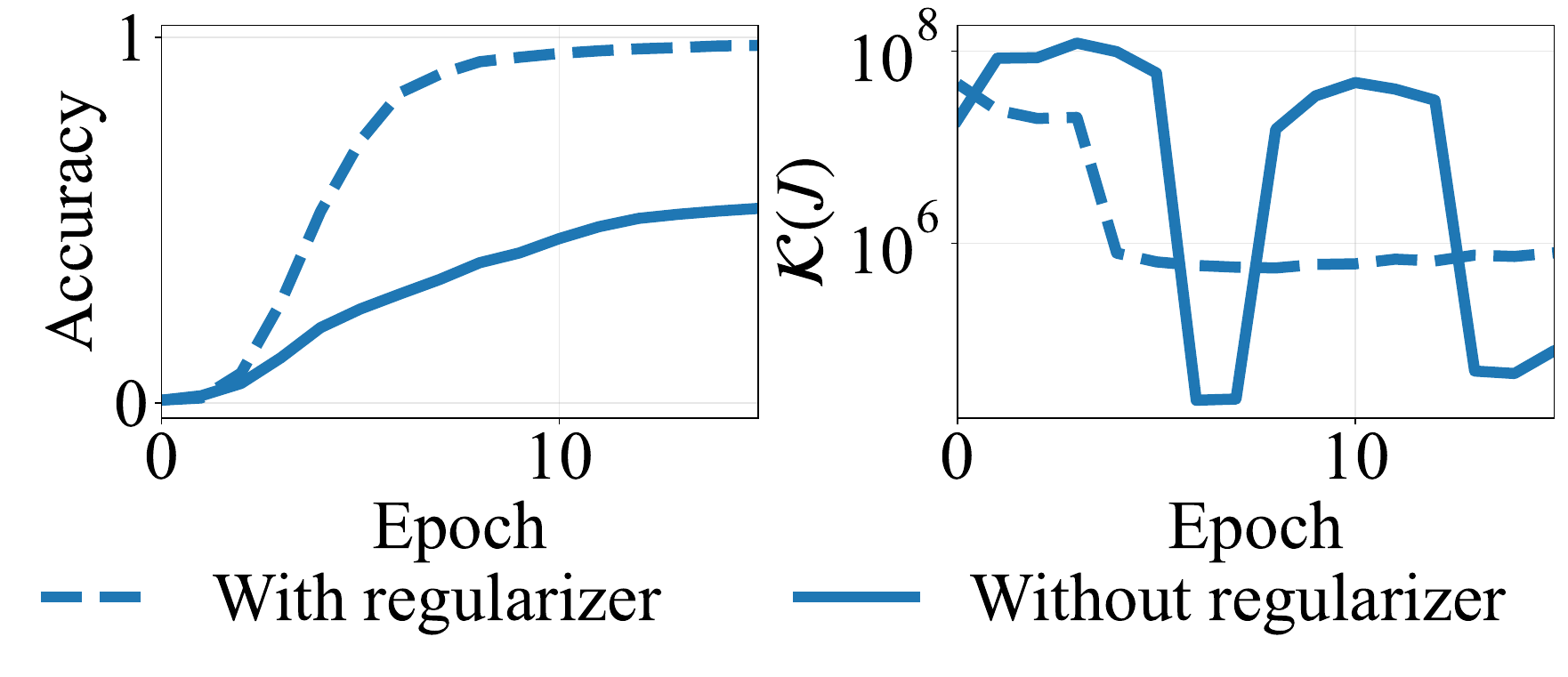}
    \caption{
        Well-tuned training dynamics with and without Jacobian regularization.
        Without regularization, $\kappa(J)$ fluctuates to extremes, resulting in performance stagnation. 
        In contrast, Jacobian regularization
        stabilizes $\kappa(J)$ and leads to maximum performance.
    }
    \label{fig:epoch_vs_jacobian_regularizer}
\end{wrapfigure}
The results above suggest a strong connection between degraded contrastive performance and ill-conditioned encoder Jacobians. To determine whether this relationship reflects an underlying mechanism rather than a simple correlation, we perform a direct intervention on encoder geometry using the Jacobian regularizer introduced in \Cref{sec:method}. Constraining\eg the Triangle objective with local directional gains improves Jacobian conditioning throughout training, and leads to both faster convergence and higher retrieval performance exemplified on Synthetic-\texttt{XNOR} (\Cref{fig:epoch_vs_jacobian_regularizer}). These improvements mirror those obtained by \acp{gpe}. Because the intervention acts directly on encoder Jacobians rather than modifying the contrastive objective,
these results provide evidence consistent with a mechanism in which encoder Jacobian conditioning is a key factor governing multimodal contrastive learning performance.

\paragraph{LeakyReLU Analysis}
\begin{wrapfigure}[12]{r}{0.55\textwidth}
    \centering
    \includegraphics[
            width=\linewidth,
            trim=0 1.05cm 0 0,
            clip
    ]{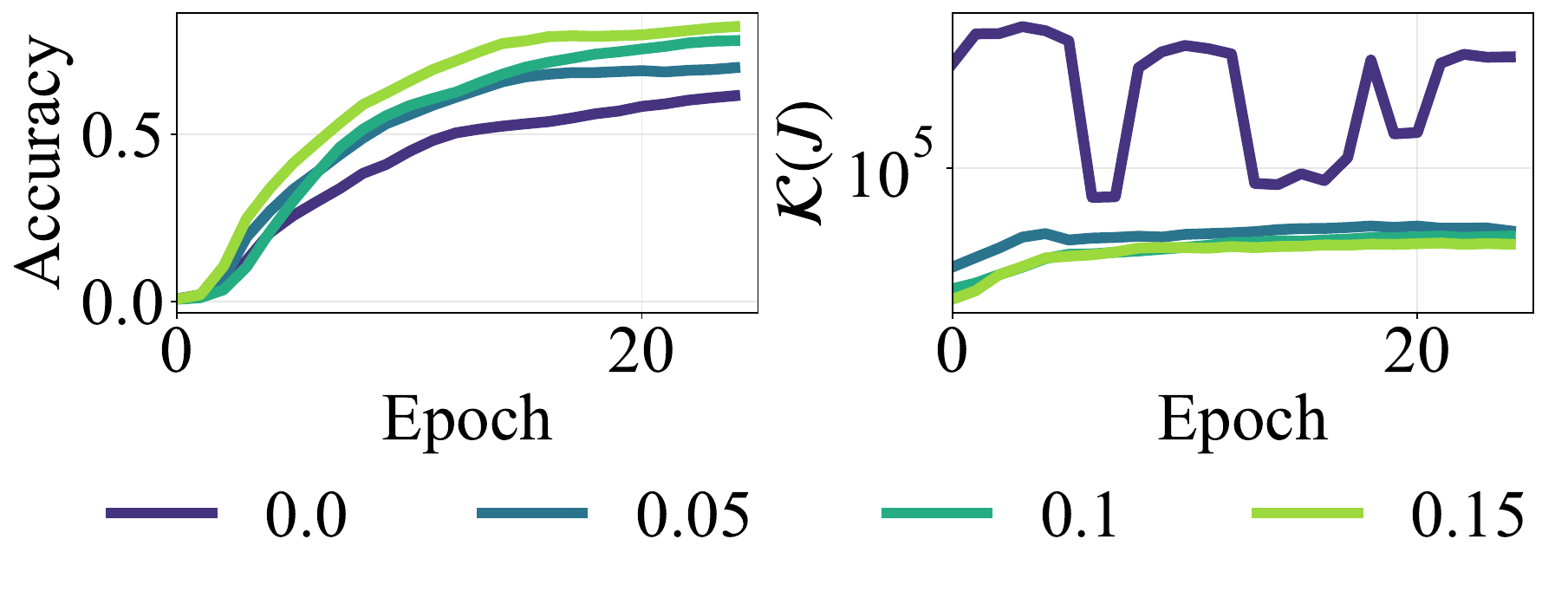}
    \caption{
        Well-tuned analysis of increasing LeakyReLU's negative slope $\alpha$
        on the Synthetic-\texttt{XNOR} dataset. Higher $\alpha$ improves accuracy by preserving non-exploding condition numbers $\kappa(J)$.
    }
    \label{fig:leakyrelu_alpha_vs_retrieval}
\end{wrapfigure}
To better understand Jacobian conditioning in case of the LeakyReLU \ac{gpe}, we vary the negative slope parameter \(\alpha\). As shown in \Cref{fig:leakyrelu_alpha_vs_retrieval}, larger values of \(\alpha\) improve retrieval performance for the Triangle objective. This trend coincides with improved Jacobian conditioning and reduced singular value collapse or explosion, supporting the hypothesis that preserving local encoder geometry is critical for effective multimodal alignment. Notably, these gains are obtained without modifying the contrastive objective itself, further emphasizing the importance of encoder geometry as an independent factor governing multimodal learning success.

\subsection{Downstream Performance}
\label{sec:downstream_performance}

\paragraph{Retrieval}
\begin{table}[t]
    \centering
    \caption{
        Comparison of well-tuned \ac{sota} multimodal contrastive objectives with and without \acp{gpe} across different datasets. Values represent top-1 accuracy of the retrieval task (mean $\pm$ \ac{se}). The results for the \ac{ukb} are highlighted in \Cref{fig:gp_vs_original}. Best non-overlapping value bold.
    }
    \label{tab:results}
    \small
    \begin{tabular}{@{}l c c c c c@{}}
    
    \toprule
    
    \textbf{Method} & 
    \textbf{Synthetic} {$\uparrow$} & 
    \textbf{\ac{ukb} $\uparrow$} &
    \textbf{\ac{ukb}-Union $\uparrow$} &
    \textbf{MIMIC-Symile $\uparrow$} &
    \textbf{MIMIC-IV $\uparrow$} \\
    
    \midrule

    Clip 
        & $0.2434$ 
        & $0.3860 \pm 0.014$
        & $0.3157 \pm 0.007$
        & $0.3075 \pm 0.008$
        & $0.6547 \pm 0.009$ \\
        
    + \ac{gpe}
        & $\mathbf{0.5814}$ 
        & $\mathbf{0.6944 \pm 0.013}$ 
        & $\mathbf{0.5448 \pm 0.013}$
        & $\mathbf{0.3604 \pm 0.009}$
        & $\mathbf{0.6985 \pm 0.003}$ \\
    \midrule

    Triangle 
        & $0.6093$ 
        & $0.5432 \pm 0.013$
        & $0.3474 \pm 0.011$
        & $0.1785 \pm 0.015$
        & $0.6226 \pm 0.008$ \\ 
        
    + \ac{gpe}
        & $\mathbf{0.9952}$ 
        & $\mathbf{0.7434 \pm 0.011}$ 
        & $\mathbf{0.5647 \pm 0.009}$
        & $\mathbf{0.3045 \pm 0.013}$
        & $\mathbf{0.7089 \pm 0.003}$ \\
    \midrule

    Gram 
        & $0.4864$ 
        & $0.1498 \pm 0.018$
        & $0.2222 \pm 0.007$
        & $0.1370 \pm 0.004$
        & $\mathbf{0.6276 \pm 0.002}$ \\
        
    + \ac{gpe}
        & $\mathbf{0.9967}$ 
        & $\mathbf{0.6293 \pm 0.013}$ 
        & $\mathbf{0.4375 \pm 0.023}$
        & $\mathbf{0.2325 \pm 0.007}$
        & $0.6087 \pm 0.006$ \\
    \midrule

    Symile 
        & $0.3310$ 
        & $0.6266 \pm 0.015$
        & $0.5165 \pm 0.008$
        & $0.3943 \pm 0.007$
        & $0.6874 \pm 0.001$ \\
    + \ac{gpe}
        & $\mathbf{0.5385}$ 
        & $\mathbf{0.7267 \pm 0.012}$ 
        & $\mathbf{0.6309 \pm 0.007}$
        & $\mathbf{0.4529 \pm 0.011}$
        & $\mathbf{0.7125 \pm 0.001}$ \\
    

    \bottomrule
\end{tabular}
\end{table}
\acp{gpe} generally improve retrieval performance across multimodal contrastive objectives and datasets (\Cref{tab:results}). On Synthetic-\texttt{XNOR}, all objectives benefit from \acp{gpe}, especially higher-order objectives such as Triangle and Gram. Similar trends are observed on both \ac{ukb} cohorts, where \acp{gpe} yield consistent improvements under complete observations as well as under missing-modality conditions. On MIMIC-IV, \acp{gpe} remain beneficial despite the presence of a LoRA-adapted BERT encoder, indicating that preserving geometry within the trainable projection encoders continues to improve multimodal alignment. However, on MIMIC-IV, Gram represents an outlier and does not benefit from \acp{gpe}. On MIMIC-Symile, \acp{gpe} consistently improve retrieval across all objectives, suggesting that geometry-preservation also benefits image-, signal-, and laboratory-based alignment. The results demonstrate a general benefit of \acp{gpe} across objectives and retrieval tasks.

\paragraph{Linear Probes}
\begin{wrapfigure}[11]{r}{0.6\textwidth}
        \vspace{-.43cm}
        \centering
        \includegraphics[
            width=\linewidth,
            trim=0 0.8cm 0 0,
            clip
        ]{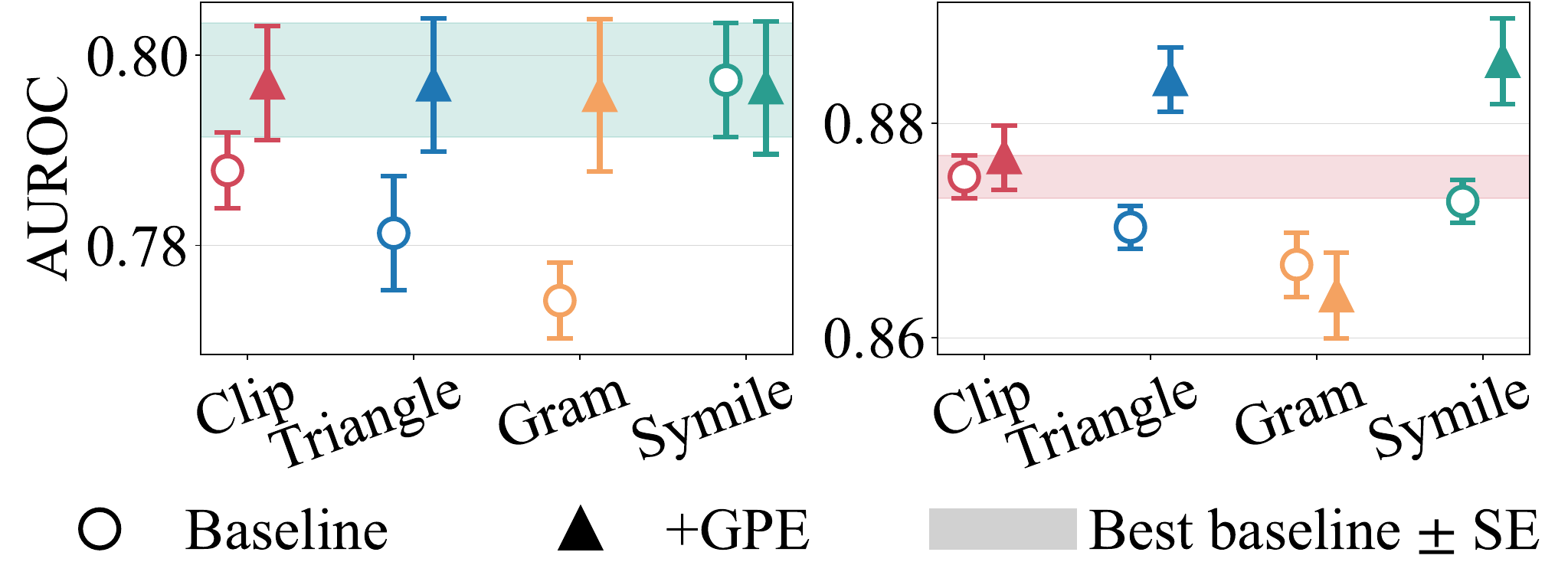}
        \caption{
            \acp{gpe} match or exceed non-\ac{gpe} linear probe baselines. Results for \ac{ukb} (left) and MIMIC-IV (right).
        }
        \label{fig:retrieval_vs_probe}
\end{wrapfigure}
We further evaluate representation quality through linear probing on mortality prediction tasks (\Cref{tab:ukb_probe_roc_auc,tab:mimic_probe_roc_auc,fig:retrieval_vs_probe}). On the \ac{ukb}, \acp{gpe} improve multimodal probe performance across CLIP, Triangle, and Gram, while maintaining competitive unimodal representations. Similar improvements are observed on MIMIC-IV, where geometry preservation generally yields higher multimodal AUROC despite more modest changes at the unimodal level. Notably, the gains observed for multimodal probes are often larger and more consistent than those observed for individual modalities, suggesting that geometry preservation primarily improves cross-modal alignment.
These findings complement the retrieval results and indicate that the improved optimization dynamics induced by \acp{gpe} translate into more useful multimodal representations for downstream prediction tasks. While objective choice can strongly affect alignment quality, representations learned by different objectives achieve comparatively similar downstream predictive performance after linear probing (\Cref{fig:retrieval_vs_probe}).
Similar observations have been reported in generative modeling, where more expressive objectives do not necessarily translate into superior linear probe performance \citep{Esmati_Nath_Hofmann_Nowrouzezahrai_Kahou_Mirmehdi_2026}. In contrast, \acp{gpe} generally improve linear probe performance across objectives and reduce the dependence of downstream representation quality on objective choice (\Cref{fig:retrieval_vs_probe}), suggesting that encoder geometry plays an important role alongside objective expressivity.
\begin{table}[t]
    \centering
    \caption{
        Well-tuned modality probes on the \ac{ukb}. Values represent the AUROC for 10 year mortality prediction (mean $\pm$ \ac{se}). Best non-overlapping value bold, best overlapping value underlined.
    }
    \label{tab:ukb_probe_roc_auc}
    \small
    \begin{tabular}{@{}l c c c c@{}}
    
    \toprule
    
    \textbf{Method} & 
    \textbf{Metabolomics $\uparrow$} &
    \textbf{EHR $\uparrow$} &
    \textbf{Proteomics $\uparrow$} &
    \textbf{Multimodal $\uparrow$} \\
    
    \midrule

    Clip
        & $\underline{0.7153} \pm 0.014$ 
        & $0.7687 \pm 0.005$ 
        & $0.7929 \pm 0.003$ 
        & $0.7879 \pm 0.004$ \\
    + \ac{gpe}
        & $0.7015 \pm 0.013$ 
        & $\underline{0.7743} \pm 0.008$ 
        & $\mathbf{0.8006} \pm 0.004$ 
        & $\underline{0.7971} \pm 0.006$ \\
    \midrule

    Triangle
        & $0.6905 \pm 0.014$ 
        & $0.7492 \pm 0.005$ 
        & $0.7806 \pm 0.003$ 
        & $0.7813 \pm 0.006$ \\
    + \ac{gpe}
        & $\underline{0.6998} \pm 0.014$ 
        & $\mathbf{0.7689} \pm 0.007$ 
        & $\mathbf{0.7949} \pm 0.005$ 
        & $\mathbf{0.7969} \pm 0.007$ \\
    \midrule

    Gram
        & $0.6566 \pm 0.015$ 
        & $0.7552 \pm 0.004$ 
        & $0.7734 \pm 0.004$ 
        & $0.7742 \pm 0.004$ \\
    + \ac{gpe}
        & $\mathbf{0.6981} \pm 0.013$ 
        & $\mathbf{0.7747} \pm 0.008$ 
        & $\mathbf{0.8039} \pm 0.002$ 
        & $\mathbf{0.7958} \pm 0.008$ \\
    \midrule

    Symile
        & $0.7042 \pm 0.014$ 
        & $0.7769 \pm 0.007$ 
        & $0.7991 \pm 0.003$ 
        & $\underline{0.7974} \pm 0.006$ \\
    + \ac{gpe}
        & $\underline{0.7048} \pm 0.014$ 
        & $\underline{0.7781} \pm 0.006$ 
        & $\underline{0.8004} \pm 0.004$ 
        & $0.7966 \pm 0.007$ \\

    \bottomrule
\end{tabular}
\end{table}

\begin{table}[t]
    \centering
    \caption{
        Well-tuned modality probes on MIMIC-IV. Values represent the AUROC for in-hospital mortality prediction (mean $\pm$ \ac{se}).
        Best non-overlapping value bold, best overlapping value underlined.
    }
    \label{tab:mimic_probe_roc_auc}
    \small
    \begin{tabular}{@{}l c c c c@{}}
    
    \toprule
    
    \textbf{Method} & 
    \textbf{Reports $\uparrow$} &
    \textbf{Timeseries $\uparrow$} &
    \textbf{Demographics $\uparrow$} &
    \textbf{Multimodal $\uparrow$} \\
    
    \midrule

    Clip
        & $\mathbf{0.7834 \pm 0.013}$ 
        & $\underline{0.8747 \pm 0.002}$ 
        & $0.5992 \pm 0.003$ 
        & $0.8750 \pm 0.002$ \\
    + \ac{gpe}
        & $0.7613 \pm 0.006$ 
        & $0.8744 \pm 0.003$ 
        & $\underline{0.6002 \pm 0.003}$ 
        & $\underline{0.8768 \pm 0.003}$ \\
    \midrule

    Triangle
        & $0.7733 \pm 0.009$ 
        & $0.8716 \pm 0.002$ 
        & $0.5991 \pm 0.004$ 
        & $0.8703 \pm 0.002$ \\
    + \ac{gpe}
        & $\underline{0.7827 \pm 0.009}$ 
        & $\mathbf{0.8768 \pm 0.002}$ 
        & $\underline{0.6005 \pm 0.003}$ 
        & $\mathbf{0.8841 \pm 0.003}$ \\
    \midrule

    Gram
        & $\mathbf{0.7735 \pm 0.013}$ 
        & $\underline{0.8685 \pm 0.003}$ 
        & $0.6005 \pm 0.003$ 
        & $\underline{0.8668 \pm 0.003}$ \\
    + \ac{gpe}
        & $0.7403 \pm 0.005$ 
        & $0.8653 \pm 0.003$ 
        & $\underline{0.6011 \pm 0.003}$ 
        & $0.8639 \pm 0.004$ \\
    \midrule

    Symile
        & $\underline{0.7827 \pm 0.009}$ 
        & $0.8718 \pm 0.002$ 
        & $0.5935 \pm 0.004$ 
        & $0.8727 \pm 0.002$ \\
    + \ac{gpe}
        & $0.7782 \pm 0.007$ 
        & $\mathbf{0.8792 \pm 0.003}$ 
        & $\mathbf{0.6011 \pm 0.003}$ 
        & $\mathbf{0.8858 \pm 0.004}$ \\

    \bottomrule
\end{tabular}
\end{table}
\section{Conclusion, Limitations \& Future Work}
\label{sec:conclusion}
We showed that multimodal contrastive learning depends strongly on encoder Jacobian conditioning. Across objectives and datasets, training is characterized by collapsing minimum singular values, exploding maximum singular values, and rapidly increasing condition numbers.
Through direct Jacobian regularization, we demonstrated that encoder conditioning stabilizes optimization. Building on this observation, we introduced \acp{gpe} based on residual paths and LeakyReLU activations. These architectural modifications improved retrieval and linear probing performance across multimodal contrastive objectives and datasets.
%
We focus on trimodal contrastive learning\ie more modalities remain important for future work.
Although we evaluate both structured and unstructured modalities, our experiments are centered on healthcare datasets, motivated by the availability of diverse modalities in this domain. 
Finally, our analysis focuses on \ac{mlp}-based projections, which remain common in multimodal learning. In contrast, other architectures already incorporate architectural mechanisms that facilitate information and gradient transport and appear more robust to the Jacobian degeneration. We therefore provide a preliminary analysis of the mechanisms (\Cref{app:transformers}), leaving a more detailed investigation to future work.
More broadly, multimodal contrastive learning may benefit from a shift from increasingly expressive objectives toward a deeper understanding of encoder geometry, optimization dynamics, and information transport across modalities.



\clearpage 
\bibliography{iclr2026_conference}
\bibliographystyle{iclr2026_conference}

\clearpage 
\newpage 
\appendix
\section{Acknowledgement}
The authors acknowledge the Scientific Computing of the IT Division at the Charité Universitätsmedizin Berlin for providing computational resources that have contributed to the research results reported in this paper. This research has been conducted using the UK Biobank Resource under application number 49966.

\section{Broader Impact and Ethics}
This work studies the optimization and geometry of multimodal contrastive learning systems. Our primary contribution is a better understanding of how encoder Jacobian conditioning influences multimodal alignment and how simple architectural modifications can improve training stability. As such, the proposed methods are general-purpose and are not tied to a specific application domain.
The datasets considered in this work are either synthetic or derived from existing biomedical resources that have been extensively used for machine learning research. We do not introduce new sensitive data sources, nor do we propose methods intended for clinical decision-making without further validation. While improved multimodal representation learning may ultimately benefit downstream healthcare applications, any deployment in high-stakes settings would require rigorous evaluation for fairness, robustness, privacy, and clinical safety.
Overall, we believe the primary impact of this work is methodological: improving the reliability, stability, and scientific understanding of multimodal contrastive learning systems.

\section{Residual Transport and Singular-Value Perturbation}
\label{app:residual_theorem}
The residual transport analysis in \Cref{sec:method} relies on standard singular-value perturbation bounds. Let
\begin{equation}
    J_E = P + H,
\end{equation}
where \(P\) denotes the residual projection and \(H\) the Jacobian of the nonlinear branch. Weyl's inequality for singular values implies
\begin{equation}
    |\sigma_k(P+H)-\sigma_k(P)|
    \le
    \|H\|_2
\end{equation}
for all singular values \(\sigma_k\). Applying this result to the smallest and largest singular values yields
\begin{equation}
    \sigma_{\min}(J_E)
    \ge
    \sigma_{\min}(P)-\|H\|_2
\end{equation}
and
\begin{equation}
    \sigma_{\max}(J_E)
    \le
    \sigma_{\max}(P)+\|H\|_2.
\end{equation}
Consequently, if
\begin{equation}
    \|H\|_2 < \sigma_{\min}(P),
\end{equation}
then
\begin{equation}
    \sigma_{\min}(J_E)>0,
\end{equation}
showing that the residual projection preserves locally non-degenerate directions despite perturbations introduced by the nonlinear branch.

\section{Additional Details}
\label{app:additional_details}
In the following, we provide additional details for our experiments, setups, and results. 

\begin{figure}[h]
    \centering
    \includegraphics[width=\linewidth]{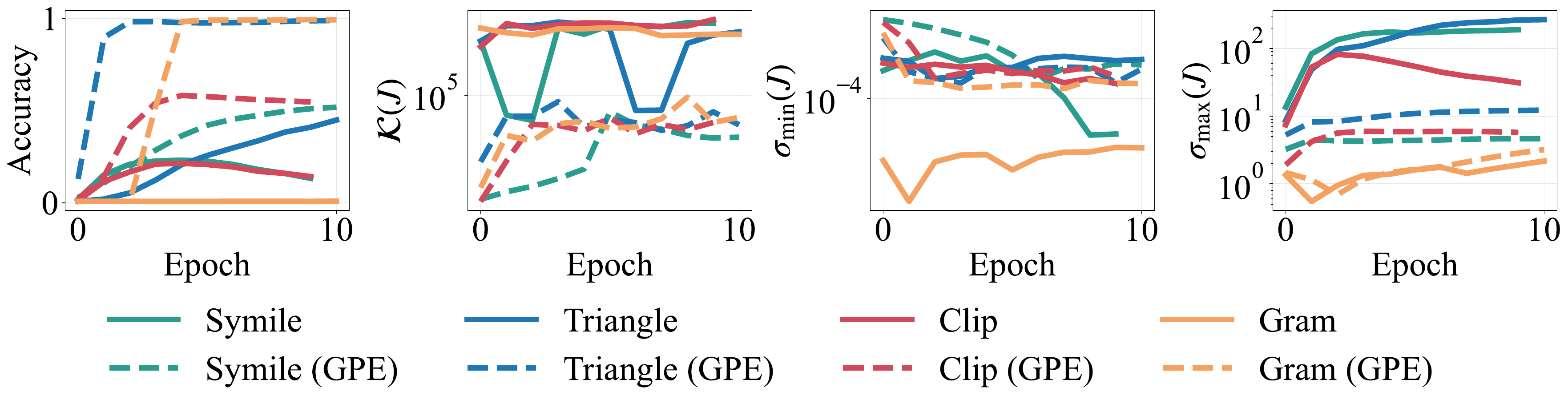}
    \caption{
        Training dynamics of standard encoders and \acp{gpe} with different scorers on the Synthetic-\texttt{XNOR} dataset. Full visualization from focused view in \Cref{fig:epoch_vs_jacobian}.
    }
    \label{fig:synth_xnor_epoch_vs_jacobian}
\end{figure}

\begin{figure}[h]
    \centering
    \includegraphics[width=\linewidth]{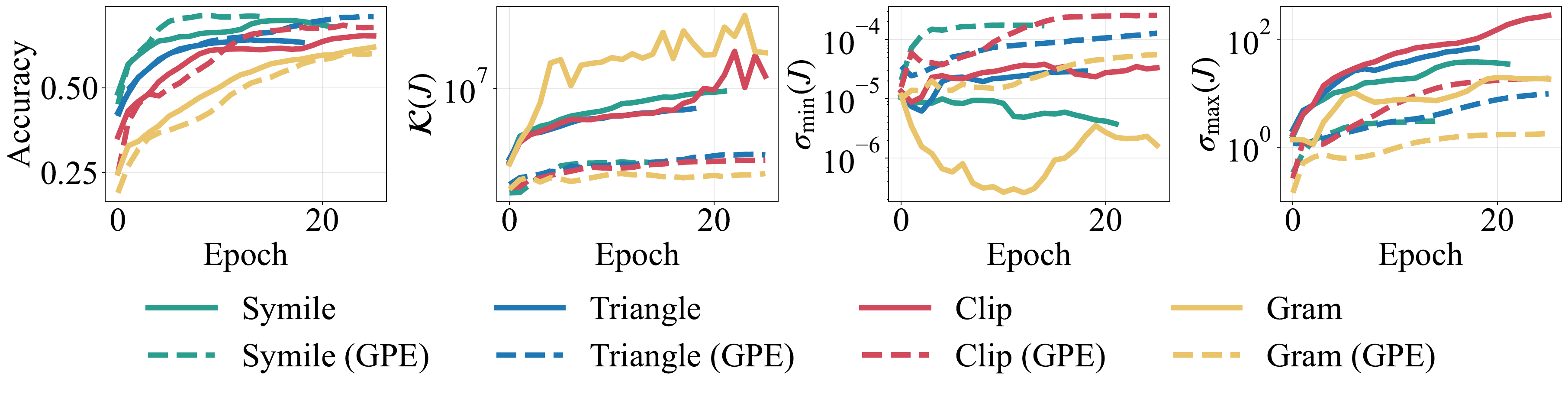}
    \caption{
        Training dynamics of standard encoders and \acp{gpe} with different scorers on the MIMIC-IV dataset. 
    }
    \label{fig:mimic_iv_epoch_vs_jacobian}
\end{figure}

\begin{figure}[h]
    \centering
    \includegraphics[width=\linewidth]{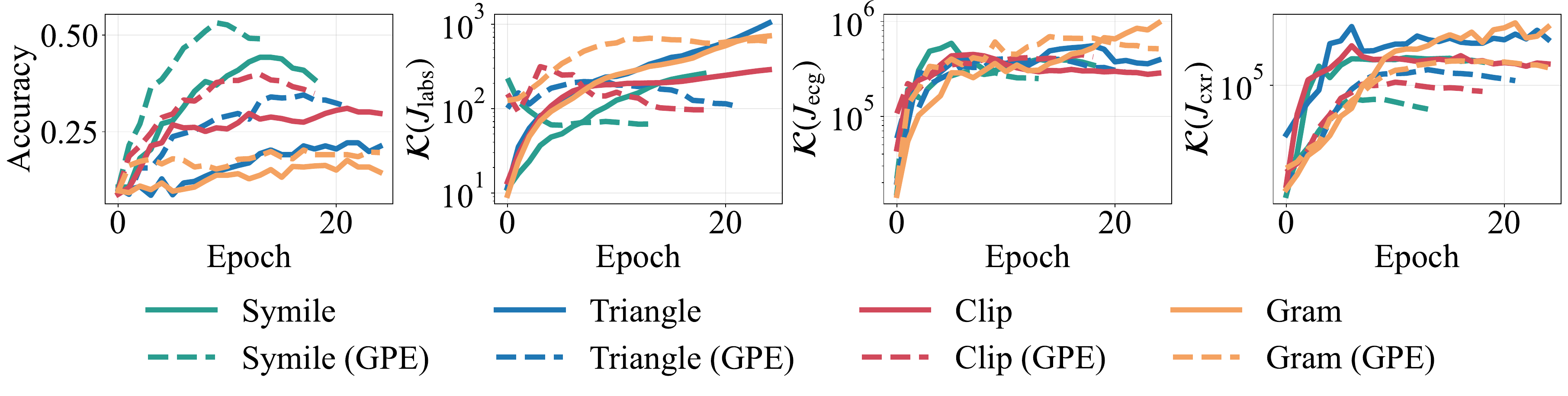}
    \caption{
        Training dynamics of standard encoders and \acp{gpe} with different scorers on the MIMIC-Symile dataset for each modality. For \acp{gpe}, labs are encoded by residaul \acp{mlp}, and ECGs and chest X-rays are encoded by ResNets. For non-\ac{gpe}, we use traditional \acp{mlp} and removed the skip connections from the ResNets.
    }
    \label{fig:mimic_symile_epoch_vs_jacobian}
\end{figure}

\paragraph{Training Dynamics} 
The full scorer training dynamics on the Synthetic-\texttt{XNOR} dataset from \Cref{fig:epoch_vs_jacobian} is visualized in \Cref{fig:synth_xnor_epoch_vs_jacobian}. The training dynamics analyses for MIMIC-IV and MIMIC-Symile are visualized in \Cref{fig:mimic_iv_epoch_vs_jacobian,fig:mimic_symile_epoch_vs_jacobian}. All Jacobian statistics are based on the encoders, with raw inputs or, in case of MIMIC-IV, RadBERT embeddings.

\paragraph{Modality Setup in MIMIC-IV}
\begin{wrapfigure}{r}{0.35\linewidth}
    \centering
    \vspace{-0.3em}
    \includegraphics[width=\linewidth]{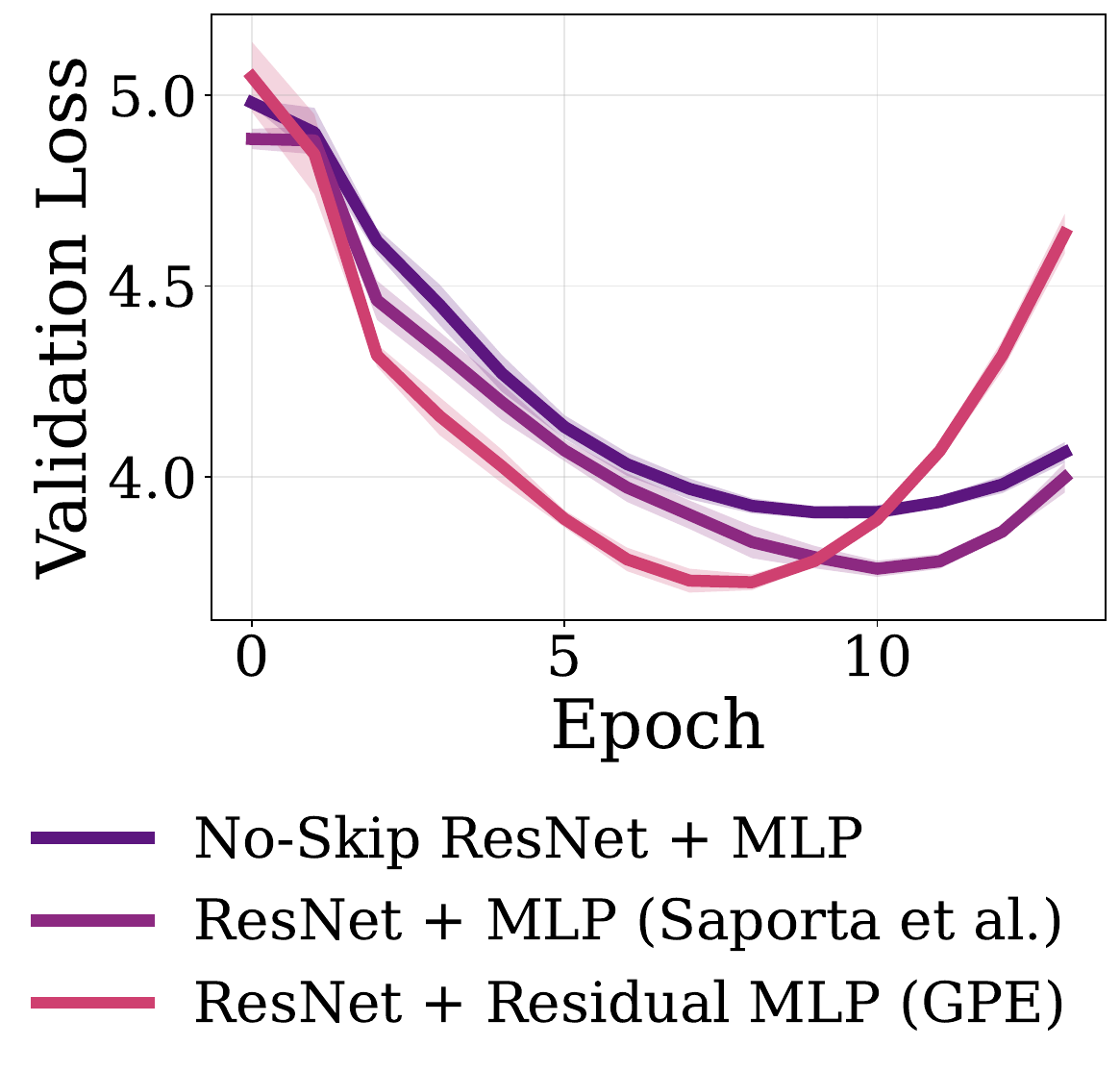}
    \caption{
        Validation loss comparison on the MIMIC-Symile dataset for the originally published encoders, non-\ac{gpe}, and \acp{gpe}. \acp{gpe} demonstrate faster convergence and lower validation loss values.
    }
    \label{fig:mimic_symile_loss_sd_comparison}
    \vspace{-1em}
\end{wrapfigure}
For our MIMIC-IV trimodal contrastive learning setup, we use three modalities per ICU stay: radiology text, clinical chart timeseries, and demographics. The radiology modality is restricted to notes with charttime in the interval from the first 48 hours, with notes after discharge excluded and notes after death excluded. Among valid notes, we keep the earliest report per stay. 
The clinical timeseries modality is restricted to measurements in the same interval, with observations after discharge removed and observations after death removed. We use the following 32 charted variables (item IDs) exactly: 220045 (Heart Rate), 220050 (Arterial Blood Pressure systolic), 220051 (Arterial Blood Pressure diastolic), 220052 (Arterial Blood Pressure mean), 220074 (Central Venous Pressure), 220179 (Non Invasive Blood Pressure systolic), 220180 (Non Invasive Blood Pressure diastolic), 220181 (Non Invasive Blood Pressure mean), 220210 (Respiratory Rate), 220277 (O$_2$ saturation pulseoxymetry), 220739 (GCS Eye Opening), 223761 (Temperature Fahrenheit), 223900 (GCS Verbal Response), 223901 (GCS Motor Response), 224054 (Braden Sensory Perception), 224055 (Braden Moisture), 224056 (Braden Activity), 224057 (Braden Mobility), 224058 (Braden Nutrition), 224059 (Braden Friction/Shear), 224168 (Parameters Checked), 224641 (Alarms On), 225664 (Glucose finger stick), 228096 (Richmond-RAS Scale), 228299 (Goal Richmond-RAS Scale), 228305 (ST Segment Monitoring On), 228409 (Strength L Arm), 228410 (Strength L Leg), 228411 (Strength R Leg), 228412 (Strength R Arm), 229321 (Activity/Mobility, JH-HLM), and 229381 (Orientation). For each item ID and ICU stay, we compute six summary statistics similar to \cite{Saporta_Puli_Goldstein_Ranganath_2024}: count, mean, std, min, max, and last. Concatenating these statistics yields a fixed-length vector of \(32 \times 6 = 192\) features. 
The demographic modality uses structured patient-level covariates\eg age and sex.

\paragraph{MIMIC-Symile Comparisons} 
\Cref{fig:mimic_symile_loss_sd_comparison} compares validation loss trajectories across CV folds and independent seeds on MIMIC-Symile. Compared with the originally published encoders (\ac{mlp} and ResNets) and the non-\ac{gpe} (\ac{mlp} and non-skip ResNets) variants, \acp{gpe} (residual \ac{mlp} and ResNets) converge more rapidly and reach lower validation losses.

\paragraph{Linear Encoder Ablation}
\begin{wraptable}[15]{r}{0.38\textwidth}
    \centering
    \vspace{-13pt}
    \caption{
        Linear encoder ablation of well-tuned objectives on the \ac{ukb}. Best non-overlapping top-1 accuracy bold and best overlapping underlined.
    }
    \label{tab:ablation_linear}
    \small
    \begin{tabular}{@{}l c@{}}
        \toprule
        \textbf{Ablation} &
        \textbf{Top-1 Accuracy} {$\uparrow$} \\
        
        \midrule
        
        Clip w/ \ac{gpe} & $\mathbf{0.6944 \pm 0.013}$ \\
        Clip w/ Linear & $0.5969 \pm 0.015$ \\
        \midrule
        
        Triangle w/ \ac{gpe} & $0.7434 \pm 0.011$ \\
        Triangle w/ Linear & $\underline{0.7438 \pm 0.012}$ \\
        \midrule

        Gram w/ \ac{gpe} & $\underline{0.6293 \pm 0.013}$ \\
        Gram w/ Linear & $0.6116 \pm 0.015$ \\ 
        \midrule

        Symile w/ \ac{gpe} & $\mathbf{0.7267 \pm 0.012}$ \\
        Symile w/ Linear & $0.3111 \pm 0.010$ \\
        \bottomrule
    \end{tabular}
\end{wraptable}
\Cref{tab:ablation_linear} compares \acp{gpe} against linear modality encoders on the \ac{ukb}. Linear encoders perform competitively for several objectives and even match Triangle with \acp{gpe}, indicating that a substantial part of the multimodal retrieval signal is accessible through stable low-complexity projections. 
However, the strong degradation for Symile and the lower performance for Clip and Gram show that linear encoders are not uniformly sufficient, suggesting that \acp{gpe} can retain the stability of simple projections while preserving objective-dependent nonlinear flexibility. 
Moreover, our MIMIC-Symile results show that \acp{gpe} also transfer to settings with intrinsically nonlinear modalities such as chest X-ray images and \acp{ecg}.

\section{Transformers Contain Geometry-Preserving Mechanisms}
\label{app:transformers}
\begin{wraptable}[16]{r}{0.65\textwidth}
    \centering
    \vspace{-9pt}
    \caption{
        Effect of token mixing and residual transport on convergence for Transformer encoders. Learned self-attention is not strictly necessary: fixed uniform token mixing also converges when residual transport is present. In contrast, removing both residual transport and token mixing causes optimization collapse. These results suggest that dense transport paths, rather than attention expressivity alone, are the primary stabilizing mechanism.
    }
    \label{tab:ablation_transformer}
    \begin{tabular}{@{}l c c c@{}}

    \toprule
    Architecture & 
    Token Mixing & 
    Residual & 
    Convergence \\
    
    \midrule
    
    Self-Attention & 
    \ding{51} & 
    \ding{51} & 
    \ding{51} \\
    Self-Attention & 
    \ding{51} & 
    \ding{55} & 
    \ding{51} \\
    Uniform Attention & 
    \ding{51} & 
    \ding{51} & 
    \ding{51} \\
    
    Uniform Attention & 
    \ding{51} & 
    \ding{55} & 
    \ding{55} \\
    Token-wise MLP & 
    \ding{55} & 
    \ding{55} & 
    \ding{55} \\
    
    \bottomrule
    
    \end{tabular}
\end{wraptable}
In the main paper, we observed that geometry-preserving interventions are particularly beneficial for modality encoders implemented as \acp{mlp} and optionally on top of LoRA-finetuned RadBERT models. In contrast, Transformer encoders exhibited stable optimization behavior even without explicitly introducing geometry-preserving modifications. To better understand this phenomenon, we investigate which architectural components are responsible for this robustness. Further, we refer to \cite{Ji_Saratchandran_Moghadam_Lucey_2025} for related work in this direction.
Since the original Synthetic-\texttt{XNOR} benchmark consists of fixed-size feature vectors, we extend the dataset to a two-dimensional sequential variant that can be processed by sequence models. We then evaluate a series of Transformer-inspired architectures that progressively remove components commonly believed to contribute to optimization stability (\Cref{tab:ablation_transformer}).
\emph{Self-Attention} corresponds to a standard Transformer encoder layer with learned query, key, and value projections. 
\emph{Uniform Attention} removes the learned attention mechanism and instead replaces each token representation by the mean of all tokens, thereby preserving token mixing while eliminating learned attention weights. 
Finally, \emph{Token-wise MLP} processes each token independently through identical feed-forward layers and therefore removes both learned attention and cross-token information exchange.
Standard self-attention Transformers converge both with and without residual transport paths. Interestingly, replacing learned self-attention by fixed uniform token mixing does not prevent convergence as long as residual transport remains present. In contrast, removing residual transport from the uniform mixing architecture leads to optimization collapse. Likewise, a token-wise MLP without token mixing and without residual transport fails to learn the task.
These findings suggest that the robustness of Transformer architectures cannot be explained solely by the expressivity of learned self-attention. Instead, the results indicate that dense information transport across tokens, together with residual transport paths, provides a strong inductive bias against the Jacobian degeneration observed in standard MLP encoders. This observation is consistent with recent work arguing that skip connections play a central role in Transformer optimization and may be more important than attention expressivity itself for stable training \citep{Ji_Saratchandran_Moghadam_Lucey_2025}.
While these experiments do not isolate every component of the Transformer architecture, they provide additional evidence for the central hypothesis of this work: successful multimodal contrastive learning depends critically on preserving trainable encoder geometry. Transformer architectures appear naturally robust to ill-conditioned Jacobians because they already contain architectural mechanisms that facilitate information and gradient transport throughout the network.

\section{Compute Environment}
\label{app:compute_environment}
We computed every experiment on a \ac{hpc} with the following environment:
\begin{enumerate}
    \item 21 Dell PowerEdge R7525 compute nodes, each with 64 AMD Epyc cores (Rome), 512GB RAM and 1 NVIDIA A100 40G GPU
    \item 2 Dell PowerEdge XE8545 compute nodes, each with 128 AMD Epyc cores (Milan), 512GB RAM, 4 NVIDIA A100 40G and 4 NVIDIA A100 80G GPUs (NVLink-connected)
\end{enumerate}

\section{Hyperparameter Tuning}
\label{app:hyperparameter_tuning}
Hyperparameters are selected using Bayesian optimization with validation loss as the optimization objective. Each method is tuned using \(100\) optimization trials. Following common practice \citep{tuningplaybookgithub}, batch size is fixed in advance and excluded from the search space.
For Synthetic-\texttt{XNOR}, hyperparameters are retuned for each experimental setting, including different values of LeakyReLU's negative slope $\alpha$. Due to the substantially larger computational requirements of \ac{ukb}-Union, the number of Bayesian optimization trials is reduced to \(50\) for this dataset. 
For the Jacobian regularizer experiments, we use $K=32$ random probe directions per sample (computed every training step) across all three modalities. We set $m=0.5$, $M=5.0$, $w_{\min}=10.0$, and $w_{\max}=1.0$. The intervention is added to the base contrastive objective as $\mathcal{L}=\mathcal{L}_{\text{base}}+\lambda \mathcal{L}_{\mathrm{JVP}}$ with $\lambda=1.0$.
The complete search spaces for all methods, linear probes, and datasets are provided in \Cref{lst:clip_triangle_gram_symile,lst:gpe,lst:mimic_iv,lst:synthetic_xnor,lst:mimic_symile,lst:ukb,lst:linear-probe-space}.

\begin{listing}[ht]
\begin{minted}
[
    framesep=2mm,
    linenos,
    fontsize=\scriptsize,
]
{yaml}
method: bayes
metric:
  name: val/min_loss
  goal: minimize

modelname.logit_scale_init:
    min: -3
    max: 0
    distribution: "uniform"
optimizer.lr:
    min: 0.00001
    max: 0.01
    distribution: "log_uniform_values"
optimizer.warmup_steps:
    values: [0, 10, 50, 100, 200, 500, 1000, 1200]
optimizer.weight_decay:
    values: [0, 0.1, 0.01, 0.001]
\end{minted}
\caption{Hyperparameters related to Clip, Triangle, Gram and Symile.}
\label{lst:clip_triangle_gram_symile}
\end{listing}
\begin{listing}[ht]
\begin{minted}
[
    framesep=2mm,
    linenos,
    fontsize=\scriptsize,
]
{yaml}
method: bayes
metric:
  name: val/min_loss
  goal: minimize

encoders.residual_path:
    values: [True]
encoders.leaky_relu_negative_slope:
    values: [0.0, 0.05, 0.1, 0.2, 0.3, 0.4, 0.5, 0.6, 0.7, 0.8]
\end{minted}
\caption{Hyperparameters related to \acp{gpe} of the objectives.}
\label{lst:gpe}
\end{listing}
\begin{listing}[ht]
\begin{minted}
[
    framesep=2mm,
    linenos,
    fontsize=\scriptsize,
]
{yaml}
method: bayes
metric:
  name: val/min_loss
  goal: minimize

modelname.emb_dim:
    values: [256]  # initially tuned from 32-1024
modelname.embedding_norm:
    values: [True]
# Encoders fixed to MLPs
\end{minted}
\caption{Hyperparameters related to Synthetic-\texttt{XNOR}.}
\label{lst:synthetic_xnor}
\end{listing}
\begin{listing}[ht]
\begin{minted}[
    framesep=2mm,
    linenos,
    fontsize=\scriptsize,
]
{yaml}
method: bayes
metric:
  name: val/min_loss
  goal: minimize

modelname.emb_dim:
    values: [1024]  # initially tuned from 256-8196
modelname.embedding_norm:
    values: [True]
# Encoders fixed to MLPs + LoRA-finetuned RadBERT
\end{minted}
\caption{Hyperparameters related to MIMIC-IV.}
\label{lst:mimic_iv}
\end{listing}
\begin{listing}[ht]
\begin{minted}[
    framesep=2mm,
    linenos,
    fontsize=\scriptsize,
]
{yaml}
method: bayes
metric:
  name: val/min_loss
  goal: minimize

modelname.emb_dim:
    values: [1024]  # initially tuned from 256-8196
modelname.embedding_norm:
    values: [True]
# Encoders fixed to MLPs + ResNets
\end{minted}
\caption{Hyperparameters related to MIMIC-Symile.}
\label{lst:mimic_symile}
\end{listing}
\begin{listing}[ht]
\begin{minted}
[
    framesep=2mm,
    linenos,
    fontsize=\scriptsize,
]
{yaml}
method: bayes
metric:
  name: val/min_loss
  goal: minimize

modelname.emb_dim:
    values: [6144]  # initially tuned from 256-8196
modelname.embedding_norm:
    values: [True]
encoders.nmr.mlp.hidden_dims:
    values: [[1024,2048,4096]]  # initially tuned with 128-4096
encoders.nmr.mlp.hidden_dropouts:
    values: [[0.2,0.2,0.2]]  # initially tuned with 0.0-0.6
encoders.ehr.mlp.hidden_dims:
    values: [[1024,2048,4096]]  # initially tuned with 128-4096
encoders.ehr.mlp.hidden_dropouts:
    values: [[0.6,0.6,0.6]]  # initially tuned with 0.0-0.6
encoders.olink.mlp.hidden_dims:
    values: [[1024,2048,4096]]  # initially tuned with 128-4096
encoders.olink.mlp.hidden_dropouts:
    values: [[0.4,0.4,0.4]]  # initially tuned with 0.0-0.6
# Encoders fixed to MLPs
\end{minted}
\caption{Hyperparameters related to the \ac{ukb}.}
\label{lst:ukb}
\end{listing}
\begin{listing}[ht]
\begin{minted}[
    framesep=2mm,
    linenos,
    fontsize=\scriptsize,
]{yaml}
method: grid
selection_metric:
  values: [roc_auc]

probe_type:
  values: [logreg_saga]
C:
  values: [1e-3, 1e-2, 1e-1, 1.0]
class_weight:
  values: [balanced]
max_iter:
  value: 1000
tol:
  value: 1e-3
\end{minted}
\caption{Hyperparameter search space for linear probes on MIMIC-IV and the \ac{ukb}.}
\label{lst:linear-probe-space}
\end{listing}

\end{document}